# Bounding-box deep calibration for high performance face detection

Shi Luo[1,2] | Xiongfei Li[1,2] | Xiaoli Zhang[1,2]

[1]College of Computer Science and Technology, Jilin University, Changchun, China

[2]Key Laboratory of Symbolic Computation and Knowledge Engineering of Ministry of Education, Jilin University, Changchun, China

**Correspondence**
Xiaoli Zhang, College of Computer Science and Technology, Jilin University, Changchun 130012, China.
Email: xiaolizhang@jlu.edu.cn

**Funding information**
The Fundamental Research Funds for the Central Universities, JLU; The Graduate Innovation Fund of Jilin University; The 'Thirteenth Five-Year Pla' Scientific Research Planning Project of Education Department of Jilin Province, Grant/Award Numbers: JKH20200678KJ, JJKH20200997KJ; The National Key Research and Development Project of China, Grant/Award Number: 2019YFC0409105; The National Natural Science Foundation of China, Grant/Award Number: 61801190; The Industrial Technology Research and Development Funds of Jilin Province, Grant/Award Number: 2019C054-3

**Abstract**
Modern convolutional neural networks (CNNs)-based face detectors have achieved tremendous strides due to large annotated datasets. However, misaligned results with high detection confidence but low localization accuracy restrict the further improvement of detection performance. In this paper, the authors first predict high confidence detection results on the training set itself. Surprisingly, a considerable part of them exist in the same misalignment problem. Then, the authors carefully examine these cases and point out that annotation misalignment is the main reason. Later, a comprehensive discussion is given for the replacement rationality between predicted and annotated bounding-boxes. Finally, the authors propose a novel Bounding-Box Deep Calibration (BDC) method to reasonably replace misaligned annotations with model predicted bounding-boxes and offer calibrated annotations for the training set. Extensive experiments on multiple detectors and two popular benchmark datasets show the effectiveness of BDC on improving models' precision and recall rate, without adding extra inference time and memory consumption. Our simple and effective method provides a general strategy for improving face detection, especially for light-weight detectors in real-time situations.

## 1 | INTRODUCTION

Face detection is a fundamental and critical step towards various downstream face applications, such as face alignment [1–3], face recognition [4–6], and facial attributes synthesis [7–10] etc. Compared with traditional methods, deep learning methods especially convolutional neural networks (CNNs) have achieved remarkable successes in a variety of computer vision tasks, ranging from image classification [11–14] to object detection [15–20], which also inspire face detection. In recent years, there is a rising demand for efficient face detectors with high performance. Benefited from large annotated datasets, CNN-based face detectors have been improved significantly in the past few years. During training, they optimise detection models by reducing face classification and bounding-box regression losses in a supervised learning manner. During inference, these well-trained models are applied to predict classification scores and bounding-box coordinates simultaneously.

Despite remarkable progress, detection results with high confidence scores may not always obtain satisfied localization accuracy. Figure 1 shows some misaligned detection results on a WIDER FACE [21] validation set. Although successfully detected with high confidence scores, the localization accuracy, quantified as the IoU between predicted and annotated bounding-boxes, is still lower than expected. Each column of Figure 1 faces the similar variation in unconstrained scenarios. However, the annotated bounding-boxes do not always follow the consistent annotation policy. Instead, predicted bounding-boxes seem to be more reliable. More importantly, we doubt that the same misaligned problem also exists in the training phase, which may limits the performance of detection models.

Actually, a similar annotation problem has been extensively studied. Representatively, Geng et al. propose a label

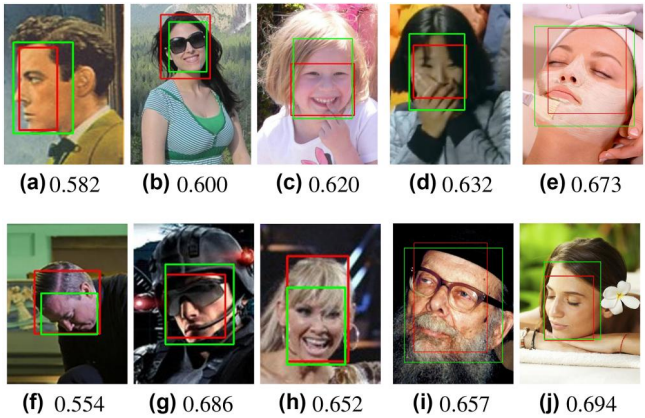

**FIGURE 1** Some misaligned results with high detection confidence but low localization accuracy on the WIDER FACE validation set. Red bounding-boxes are the face annotations. Green bounding-boxes represent the detection results. The IoUs between predicted and annotated bounding-boxes are noted below

distribution learning (LDL) approach for age estimation [22] and head pose estimation [23]. Besides, Gao et al. [24] convert the label of each instance into a discrete label distribution and learn the label distribution by minimising a Kullback–Leibler divergence between the predicted and ground-truth label distributions using deep ConvNets. These methods jointly inspire us to solve the annotation problem in face detection.

In this paper, we first perform face detection on the training set itself. A large number of high confidence detection results suffer from the misaligned problem. Then, we carefully analyse these cases and point out that annotation misalignment is the main reason. Compared with manual annotations, predicted bounding-boxes are more suitable as ground-truth annotations of these misaligned faces. Later, a comprehensive discussion is given for the replacement rationality between predicted and annotated bounding-boxes. Finally, we propose a new Bounding-Box Deep Calibration (BDC) method to reasonably calibrate misaligned annotations of the training set. Specifically, our proposed BDC method can recognise misaligned annotations and replace them with deep model predicted bounding-boxes. The calibrated annotation file is helpful for training high performance detectors. Extensive experiments on multiple detectors and two popular benchmark datasets such as WIDER FACE and FDDB [25] show the effectiveness of BDC on improving models' precision and recall rate.

Eventually, our solution for training high performance detectors from misaligned annotations consists of three stages: generating detection results on the training set, calibrating misaligned annotations with the BDC method, and training detectors from calibrated annotations. More importantly, our BDC method merely offers calibrated annotations for the training stage. No extra inference time as well as memory consumption is added. Therefore, the proposed BDC method can directly improve the detection performance, especially for light-weight detectors in real-time situations. To our knowledge, it is the first time for improving face detection to calibrate misaligned bounding-boxes on the training set. The source code will be released.[1]

For clarity, the main contributions of this work can be summarised as follows.

- We point out that annotation misalignment on the training set is the main reason for detection results with high confidence scores but low localization accuracy.
- We propose a novel Bounding-Box Deep Calibration (BDC) method to calibrate the low-quality annotations with deep model predicted bounding-boxes.
- Our simple and effective method provides a general strategy for improving face detection without adding extra inference time and memory consumption, especially for light-weight detectors in real-time situations.
- Extensive experiments on multiple detectors and two popular benchmark datasets show the effectiveness of BDC method on improving models' precision and recall rate.

The remainder of the paper is organised as follows. Section 2 briefly reviews the related work in face detection. Section 3 examines misaligned cases and presents our proposed method. Section 4 shows our experimental results. Section 5 concludes this paper.

## 2 | RELATED WORKS

CNN-based face detection methods have attracted much attention due to their detection accuracy as well as inference efficiency. To further improve detection performance, great efforts have been made in different aspects.

### 2.1 | Network design

Ingenious network design can help detectors extract more representative face features. MTCNN [26] jointly solves face detection and alignment using several multi-task CNNs. HR [27] constructs multi-level image pyramids to boost the performance on extreme scale variations. SFA [28] presents multi-branch face detection architecture which pays more attention to faces with a small scale. FANet [29] creates a new hierarchical effective feature pyramid with rich semantics at all scales. BFBox [30] designs a FPN-attention module to joint search the face-appropriate space of backbone and FPN. HLA-Face [31] builds a joint high-level and low-level adaptation framework for dark face detection.

### 2.2 | Anchor setting

Dense anchors can improve the recall rate at the cost of adding inference time. Therefore, efficient anchor setting is still an open problem. ZCC [32] introduces a novel anchor design to guarantee high overlaps between faces and anchor boxes. PyramidBox [33] formulates a data-anchor-sampling strategy to increase the proportion of small faces in the training data. FA-RPN [34] proposes an efficient anchor placement strategy to reduce the number of anchors to detect faces.

---
[1] https://github.com/shiluo1990/BDC

## 2.3 | Anchor matching and compensation

Generating diverse training samples is a foundation of training high performance detection models. The standard anchor matching strategy [17] assigns positive and negative labels on each anchor according to the IoU threshold. S$^3$FD [35] proposes scale compensation anchor matching strategy which helps the outer faces match more anchors. DSFD [36] offers an improved anchor matching method to provide better initialisation for the regressor. HAMBox [37] helps outer faces compensate high-quality anchors, which can obtain high IoU regression bounding boxes.

## 2.4 | Training sample selection

Training samples rebalanced is essential to boost detection performance. OHEM [38] automatically selects hard examples to make training more effective and efficient. SRN [39] introduces a selective two-step classification to ignore training easy sample anchors in the second stage. Group sampling [40] emphasises the importance of balanced training samples, including both positive and negative ones, at different scales.

## 2.5 | Feature enhancement

Receptive field expansion can further enhance face features. SSH [41] adds large filters on each detection module to merge the context information. DSFD introduces a feature enhance module to enhance original features to make them more discriminable and robust. RefineFace [42] constructs an RFE module to provide more diverse receptive fields for detecting extreme-pose faces.

## 2.6 | Loss function

Excellent loss functions are convenient for model convergence. UnitBox [43] presents an IoU loss to directly regress the bounding box. Focal Loss [44] down-weights the loss assigned to well-classified examples. DIoU [45] incorporates the normalised distance between the predicted box and the target box.

Recently, researchers gradually realize that it is unreasonable to directly take the classification score as the detection confidence. The low correlation between the classification confidence and localization accuracy hurts the models' detection performance severely.

## 2.7 | Consistent detection

Inspired by IoU-aware [46], TinaFace [47] adds a parallel head with a regression head to predict the IoU between the detected and annotated bounding-box. During inference, the final detection confidence is computed by the production of original classification score and predicted IoU as formulated in Equation (1).

$$S_j = p_j^\alpha IoU_j^{1-\alpha} \qquad (1)$$

where $p_j^\alpha$ and $IoU_j^{1-\alpha}$ are the original classification score and predicted IoU of the $j$th detected box, and $\alpha \in [0, 1]$ is the parameter to control the contribution to the final detection confidence. CRFace [48] proposes a confidence ranking network with a pairwise ranking loss to re-rank the predicted confidences locally within the same image. Both of them focus on calibrating final detection confidence with localization accuracy, quantified as the IoU between predicted and annotated bounding-boxes.

Despite being extensively studied, misaligned annotations on the training set, which can also lead to the low correlation problem, is still neglected. In this paper, we propose the BDC method to recognise misaligned annotations and replace them with deep model predicted bounding-boxes.

## 3 | PROPOSED METHOD

The pipeline of our solution for training high performance detectors from misaligned annotations is illustrated in Figure 2. It can be described as three stages: (1) Prediction: We first utilise a predictor to obtain detection results on the training set itself. A high performance detector is needed for prediction stage. (2) Calibration: Then, calibrated annotations are generated by our proposed BDC method. The core idea of BDC is to recognise misaligned annotations and replace them with deep model predicted bounding-boxes. (3) Training: Finally, we combine training set images and calibrated annotations to train high performance detectors. It should be noticed that the training detector is not required to be the same as the predictor.

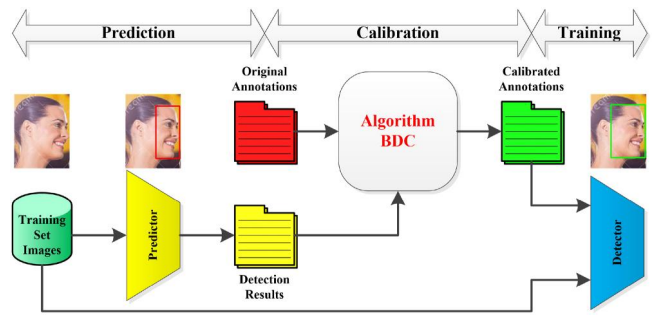

**FIGURE 2** The pipeline of our solution for training high performance face detectors from misaligned annotations. We first utilise a predictor to obtain detection results on the training set itself. Then, calibrated annotations are generated by our proposed Bounding-Box Deep Calibration (BDC) method. Finally, we adopt calibrated annotations to train high performance face detectors

## 3.1 | Misalignment category

For the purpose of recognising misaligned annotations, we first conduct face detection on the training set itself to generate detection results. Here, a simplified version of TinaFace[2] is applied as the high performance predictor. Then, we define Average Detection Confidence (ADC) to quantify high confidence detection result (HCDR).

**Definition 1** (Average Detection Confidence). Given an annotated dataset $D_{anno}$ with $N$ images, $K_a^i$ represents the number of face annotations $\hat{A}^i$ in the $i$th image. For a face detection method $M_{face}$, $R_j^i$ is the detection result, whereas $B_j^i$ represents the predicted bounding-box and $S_j^i$ is the $j$th higher detection confidence. Finally, the average detection confidence $\overline{S}_{det}$ between the face detection method $M_{face}$ and annotated dataset $D_{anno}$ is defined as

$$\overline{S}_{det}(M_{face}, D_{anno}) = \frac{\sum_{i=1}^{N}\sum_{j=1}^{K_a^i} S_j^i}{\sum_{i=1}^{N} K_a^i} \quad (2)$$

Therefore, the detection results are considered as HCDRs when their detection confidence exceed ADC $\overline{S}_{det}$.

Later, we calculate the localization accuracy, quantified as the IoU between predicted and annotated bounding-boxes, of those HCDRs. Unfortunately, many HCDRs on the training set also suffer from low localization accuracy. Table 1 lists the distribution of localization accuracy for HCDRs on the WIDER FACE training set. Although 87,476 detection results obtain high confidence scores, the localization accuracy of about 26.271% HCDRs (22,981 of all 87,476), called Misaligned Detection Result (MDR), still ranges from 0.5 to 0.8. These MDRs indicate that a similar misalignment problem is still existing in the training phase and limits the further improvement of detection performance.

Finally, we can utilise these MDRs to recognise their corresponding misaligned annotations. For convenience, we define Misaligned Bounding-Box Pair (MBP) to describe the relationship between MDRs and their best matching misaligned annotations as follows:

**Definition 2** (Misaligned Bounding-Box Pair). Given an annotated dataset $D_{anno}$ with $N$ images, $\hat{A}_k^i$ represents the $k$th face annotation of the $i$th image and $\hat{B}_k^i$ is its annotated bounding-box. For a face detectiom method $M_{face}$, $R_j^i$ is the predicted detection result, whereas $B_j^i$ represents predicted

**TABLE 1** The distribution of localization accuracy for high confidence detection results on the WIDER FACE training set

| Index | Interval | Number | Percentage (%) |
|---|---|---|---|
| 1 | [0.5,0.6] | 854 | 0.976 |
| 2 | [0.6,0.7] | 4,280 | 4.893 |
| 3 | [0.7,0.8] | 17,940 | 20.508 |
| 4 | [0.8,0.9] | 41,107 | 46.992 |
| 5 | [0.9,1.0] | 24,287 | 27.764 |
| 6 | [0.5,0.8] | 22,981 | 26.271 |
| 7 | [0.5,1.0] | 87,476 | 100.000 |

*Note*: * Detection results are predicted by TinaFace.

bounding-box best matching with $\hat{B}_k^i$ and $S_j^i$ being the $j$th higher detection confidence. Finally, we denote $MBP\left(B_j^i, \hat{B}_k^i\right)$ as misaligned bounding-box pair, which should satisfy the two conditions below:

$$MBP\left(B_j^i, \hat{B}_k^i\right) \quad s.t.$$
$$(1)\ S_j^i > \overline{S}_{det}(M_{face}, D_{anno}), \quad (3)$$
$$(2)\ IoU\left(B_j^i, \hat{B}_k^i\right) \in [T_m, T_a].$$

where $T_m$ represents the matching threshold and $T_a$ is the alignment threshold for localization accuracy. Consequently, $R_j^i$ is the MDR. Similarly, $\hat{A}_k^i$ is called misaligned annotation and $\hat{B}_k^i$ is its misaligned bounding-box, which locates the misaligned face $F_k^i$.

## 3.2 | MBP visualisation

In this section, we analyse the MBPs to seek the main reason for detection misalignment on the training set. To visualise MBPs, the predicted and annotated bounding-boxes are drawn on their corresponding training images. We observe these images and notice that many annotated faces can be successfully detected with high confidence scores.

Intuitively, the detection results with high confidence scores should be accurately located because of their rich face features. However, their localization accuracy does not always follow our expectations. Figure 3 shows some MBPs with high detection confidence scores but low localization accuracy on the WIDER FACE training set. As seen in the first row of Figure 3, both the predicted and annotated bounding-boxes are reasonable from human perspective. While low localization accuracy inevitably results in the increase of bounding-box regression loss. Compared with manual annotations on the training set, model predicted bounding-boxes are more proper in the second row of Figure 3. More importantly, few incorrect annotations still exist as shown in the last row of Figure 3. It is a wrong regression direction to drive predicted bounding-boxes to incorrect

[2] https://github.com/Media-Smart/vedadet/tree/main/configs/trainval/tinaface

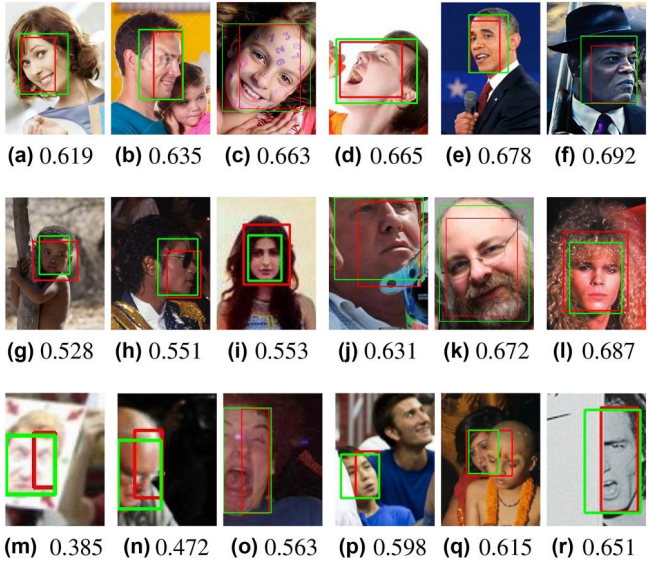

(a) 0.619 (b) 0.635 (c) 0.663 (d) 0.665 (e) 0.678 (f) 0.692
(g) 0.528 (h) 0.551 (i) 0.553 (j) 0.631 (k) 0.672 (l) 0.687
(m) 0.385 (n) 0.472 (o) 0.563 (p) 0.598 (q) 0.615 (r) 0.651

**FIGURE 3** Some misaligned bounding-box pairs on the WIDER FACE training set. Red bounding-boxes are the face annotations. Green bounding-boxes represent the detection results. The IoUs between model predicted and manual annotated bounding-boxes are noted blow

annotations. Surprisingly, predicted bounding-boxes of these incorrectly annotated faces are more fit for ground truth. Therefore, annotation misalignment on the training set is the main reason for detection results with high confidence scores but low localization accuracy. Inspired by the above analysis, we attempt to replace misaligned annotations with predicted bounding-boxes to reasonably decrease unnecessary bounding-box regression loss.

## 3.3 | Replacement rationality

We will discuss the replacement rationality from the loss function perspective. CNN-based face detectors optimise model parameters by minimising the following multi-task loss:

$$L_{total} = L_{cls} + \lambda_1 \cdot L_{reg} + \lambda_2 \cdot L_{aux} \quad (4)$$

$$L_{cls} = \sum_t \frac{1}{N_t^c} \sum_{j \in A_t} f_{cls}(\Delta p_j, \hat{p}_k) \quad (5)$$

$$L_{reg} = \sum_t \frac{1}{N_t^r} \sum_{j \in A_t} \hat{p}_k \cdot f_{reg}(\Delta q_j, \hat{q}_k) \quad (6)$$

The total loss $L_{total}$ is the weighted sum of face classification loss $L_{cls}$, bounding-box regression loss $L_{reg}$ and auxiliary loss $L_{aux}$. $L_{aux}$ is the auxiliary loss like facial landmark regression loss in MTCNN, RetinaFace and IoU prediction loss in TinaFace. $\lambda_1, \lambda_2 \in [0, 1]$ are the parameters to balance the contribution of these three subtask losses. The index $t$ goes over the detection heads $\{H_t\}$, which corresponding to multi-scale feature maps, and $A_t$ represents the set of anchors in $H_t$. $f_{cls}()$ is the face classification loss function, where $\Delta p_j$ is the predicted classification confidence of anchor $j$ being a face and $\hat{p}_k$ represents its ground-truth label (1 for positive and 0 for negative). $f_{reg}()$ is the bounding-box regression loss function, where $\Delta q_j$ is the 4 dimension coordinate parameters of the predicted bounding-box and $\hat{q}_k$ represents the ground-truth bounding-box. $N_t^c$ and $N_t^r$ are the number of anchors in detection head $H_t$, which participates in the face classification and bounding-box regression loss computation separately. It should be notice that $\hat{p}_k \cdot f_{reg}(\Delta q_j, \hat{q}_k)$ means the regression loss is activated only for positive anchors and disabled otherwise.

For each misaligned face $F_k^i$, MBP $MBP(B_j^i, \hat{B}_k^i)$ describes the relationship between misaligned detection results $R_j^i$ and its best matching misaligned annotation $\hat{A}_k^i$. So, the bounding-box regression loss for original annotation $l_{reg}^o$ can be formulated as

$$l_{reg}^o = \hat{p}_k^i \cdot f_{reg}(\Delta q_j^i, \hat{q}_k^i) \quad (7)$$

where $\Delta q_j^i$ is the predicted box of the training detector and $\hat{q}_k^i$ represents its ground-truth box sampled from original annotation $(\hat{B}_k^i, 1)$ of $F_k^i$. After performing replacement operation $Replace(\hat{B}_k^i, B_j^i)$, the bounding-box regression loss for calibrated annotation $l_{reg}^c$ can be calculated as

$$l_{reg}^c = \hat{p}_k^i \cdot f_{reg}(\Delta q_j^i, q_j^i) \quad (8)$$

where ground-truth box $q_j^i$ is sampled from calibrated annotation $(B_j^i, 1)$ of $F_k^i$.

Notice that, MDR $R_j^i$ is generated by the high performance predictor and $B_j^i$ is its predicted bounding-box. Compared with $\hat{B}_k^i$ of manual annotation $\hat{A}_k^i$, utilising $B_j^i$ as the ground-truth bounding-box of misaligned face $F_k^i$ can bring smaller bounding-box regression loss. In other words, replacing misaligned annotation with the predicted bounding-box $Replace(\hat{B}_k^i, B_j^i)$ can directly reduce unnecessary bounding-box regression loss for misaligned face $F_k^i$.

$$l_{reg}^c < l_{reg}^o \quad (9)$$

Therefore, replacement operation can reasonably decrease bounding-box regression loss $L_{reg}$ and the total loss $L_{total}$ in the training phase, which can drive the detection models to reduce the rest losses and further improve the detection performance.

## 3.4 | Bounding-box deep calibration

Face annotation methods can be roughly divided into two categories as follows. (1) Manual Annotation: We organise many people to annotate the bounding-boxes for all the recognisable faces, especially for hard detected faces. Each people should follow the annotation policy. However, some of these bounding-boxes may be inconsistent, improper, or even incorrect, called misaligned annotations. Besides, manual annotation is time-consuming and laborious. (2) Detector Prediction: When the detection methods and model parameters determined, we can rapidly predict the unique detection result for each face. Therefore, detector prediction can ensure the consistency of annotation results. However, CNN-based detectors always fail to capture hard detected faces because of their low detection confidence. These hard detected faces commonly suffer from small scale, atypical pose, extreme illumination, heavy occlusion etc. Modern annotation methods usually adopt the combination of manual annotation and detector prediction, called interactive annotation, to make best use of the advantages and bypass the disadvantages.

Inspired by modern annotation methods, we propose a new BDC method to calibrate the low-quality annotations. During the calibration stage, our BDC method can recognise misaligned annotations and replace them with deep model predicted bounding-boxes. The calibrated annotation file is offered for the following training stage. The flow chat of the BDC method can be seen in Figure 4.

Specifically, our BDC method consists of three steps: (1) Calculate the ADC score of detection results on the training set, which are generated by the high performance predictor in the prediction stage. The ADC score is used to select HCDRs from these detection results. (2) Calculate the bounding-box IoUs between HCDRs and original annotations. Some of these HCDRs are considered as MDRs when their max bounding-box IoUs belong to the calibration interval. After that, we utilise these MDRs to recognise their corresponding misaligned annotations. For convenience, we organise MBPs to describe the relationship between MDRs and their best matching misaligned annotations. (3) Create a calibration index to avoid misaligned annotations being multiple calibrated. According to the calibration index and MBPs, we update training set annotations by replacing misaligned annotations with deep model predicted bounding-boxes.

Note that, the idea of training a backbone and then using the outputs as labels to improve performance has been discussed in many fields, such as semi-supervised and unsupervised learning. These methods focus on taking model's outputs as annotations for large unlabelled data and expanding the size of the training set. Different from semi-supervised and unsupervised learning, our proposed BDC method aims at improving annotation quality of the training set. With the help of deep predictor's HCDRs, we can recognise misaligned annotations on the training set and replace them with MDRs. In short, calibrate manual annotations from deep-learning perspective. More importantly, our BDC method could be generalised to calibrate misaligned annotations with model predicted high confidence outputs in different supervised learning scenarios.

## 3.5 | Algorithm details

**Algorithm 1 Bounding-Box Deep Calibration**

**Input:** Original annotations $\hat{A}$; Detection results $R$;
**Output:** Calibrated annotations $\hat{A}_c$;
1: Calculate the ADC $\overline{S}_{det}$, as formulated in Equation (2)
2: Initialise IoU calibration interval $[T_m, T_c]$
3: **for** $i \leftarrow 1$ to $N$ **do**
4:     **for** $j \leftarrow 1$ to $K_a^i$ **do**
5:         Obtain dataset annotations $\hat{A}_j^i = (\hat{B}_j^i, 1)$
6:         Append $\hat{A}^i \leftarrow \{\hat{A}_j^i\}, \hat{B}^i \leftarrow \{\hat{B}_j^i\}$
7:     **end for**
8:     $\widetilde{K_p^i} \leftarrow 0$
9:     **for** $j \leftarrow 1$ to $K_p^i$ **do**
10:       Obtain predicted detection results $R_j^i = \left(B_j^i, S_j^i\right)$
11:       **if** $S_j^i \leq \overline{S}_{det}$ **then**
12:         break
13:       **end if**
14:       Append $B^i \leftarrow \left\{B_j^i\right\}$
15:       $\widetilde{K_p^i} \leftarrow \widetilde{K_p^i} + 1$
16:     **end for**
17:     Calculate $overlaps \leftarrow \text{IoU}\left(B^i, \hat{B}^i\right)$
18:     Obtain $max\_overlaps$ and $argmax\_overlaps$
19:     Set calibration index $C_{\text{index}} \leftarrow argmax\_overlaps[:]$
20:     $A_{\text{status}} \leftarrow \text{np.empty}\left(K_a^i\right)$
21:     $A_{\text{status}}.\text{fill}(0)$
22:     **for** $j \leftarrow 1$ to $\widetilde{K_p^i}$ **do**
23:       **if** $max\_overlaps[j] \in [T_m, T_c]$ **then**
24:         **if** $A_{\text{status}}[argmax\_overlaps[j]] == 0$ **then**
25:           $A_{\text{status}}[argmax\_overlaps[j]] \leftarrow 1$
26:           continue
27:         **end if**
28:         $C_{\text{index}}[j] \leftarrow -1$

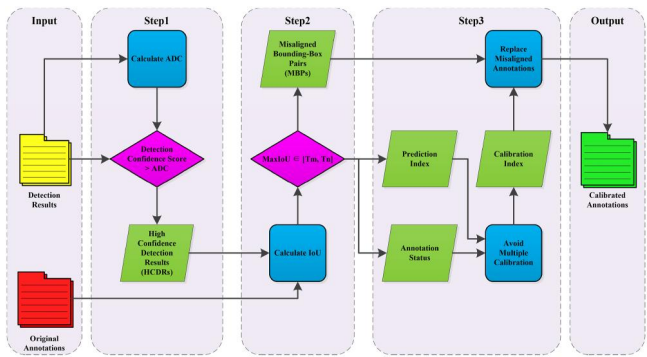

**FIGURE 4** The flow chat of our proposed Bounding-Box Deep Calibration (BDC) method

```
29:        continue
30:     end if
31:     C_index[j] ← -1
32:   end for
33:   Update Â^i ← Replace(B̂^i, B^i, C_index)
34: end for
35: return calibrated annotations Â_c ← {Â^i}
```

The details of the BDC method can be seen in Algorithm 1. First, we calculate the ADC as defined in Def.1. The IoU calibration interval $[T_m, T_c]$ is also initialled. $T_m$ represents the matching threshold while $T_c$ is for calibration. Then, we recurrently obtain the set of annotated $\hat{B}^i$ and predicted $B^i$ bounding-boxes on training set $D_{anno}$ with $N$ images, where $K_a^i$ and $K_p^i$ indicate their number, respectively. Specifically, only $\widetilde{K_p^i}$ of the total $K_p^i$ predicted detection results are HCDRs and their bounding-boxes are collected in $B^i$. For each image, we calculate the IoU between predicted $B^i$ and annotated $\hat{B}^i$ bounding-boxes. Therefore, the maximum IoU between each predicted and all annotated bounding-boxes can be noted by *max_overlaps* as well as *argmax_overlaps* for the corresponding index of the annotated bounding-box. Later, we need to determine the calibration index, denoted as $C_{index}$, according to two rules. (1) For each predicted bounding-box, the max IoU is in the calibration interval $[T_m, T_c]$. (2) The corresponding annotated bounding-box is the first time to be calibrated. In order to avoid a single annotated bounding-box being calibrated by multiple predicted ones, we adopt $A_{status}$ to mark the calibration status of all annotated bounding-boxes, where 0 stands for uncalibrated and the reverse of 1. It should be noted that these predicted bounding-boxes $B^i$ are sorted in descending order according to their detection confidence. As a result, we remove those predicted bounding-boxes, where there is no need to execute replacement, from the calibration index by labelling −1. Finally, we replace annotated bounding-boxes $\hat{B}^i$ with predicted bounding-boxes $B^i$ according to the calibration index $C_{index}$, marked as $Replace(\hat{B}^i, B^i, C_{index})$. We record the updated annotations of each image and output a new calibrated annotation file for the training set.

# 4 | EXPERIMENTS

## 4.1 | Datasets

### 4.1.1 | WIDER FACE dataset

It has 32,203 images with 393,703 annotated faces with a high degree of variability in scale, pose and occlusion. These images are divided into training (40%), validation (10%), and testing (50%) sets by randomly selecting from 61 event classes. Faces in this dataset are classified into Easy, Medium, and Hard subsets according to their detection difficulty.

### 4.1.2 | FDDB dataset

It contains 2845 images and 5171 annotated faces. Most of these faces have large scale, high resolutions or slightly occlusion sometimes. Different from WIDER FACE, faces in the FDDB dataset are annotated by bounding ellipses. In order to verify generalisation ability of our method, we perform the evaluation on the FDDB dataset.

## 4.2 | Implementation details

### 4.2.1 | Baseline

We adopt a simplified version of TinaFace as our baseline face detector. Multi-task losses are applied as the objective function. Specifically, the losses of face classification, bounding-box regression and IoU prediction are Focal Loss, DIoU Loss, and Cross-Entropy Loss, respectively.

### 4.2.2 | Calibrated annotations

We first utilise TinaFace as the predictor to generate detection results on the WIDER FACE training set. Then, we perform our proposed BDC method to replace misaligned annotations with predicted bounding-boxes. During the calibration stage, our BDC method calibrates 22,981 misaligned annotations and takes 35.66s to offer a calibrated annotation file on CPU i7-6850K. The detail parameters can be seen in Table 2.

### 4.2.3 | Training and inference settings

We follow the official settings of each detection method to train models with original or calibrated annotations, and evaluate them on the WIDER FACE validation set and FDDB dataset. All the experiments are executed on 2 NVIDIA GeForce GTX 1080Ti GPUs.

## 4.3 | Ablation study

### 4.3.1 | The effectiveness of BDC on the baseline detector

We discuss the effect of IoU calibration interval $[T_m, T_c]$ in our proposed BDC method. As seen in Figure 5, the matching IoU $T_m$ is empirically set to 0.5. Therefore, almost all annotated bounding-boxes of these overlap faces will not be merged by

**TABLE 2** The calibrated annotation file generated by our proposed BDC

| Predictor | ADC | $[T_m, T_c]$ | Calibrated | Time(s) |
|---|---|---|---|---|
| TinaFace | 0.568973 | [0.5, 0.8] | 22,981 | 35.66 |

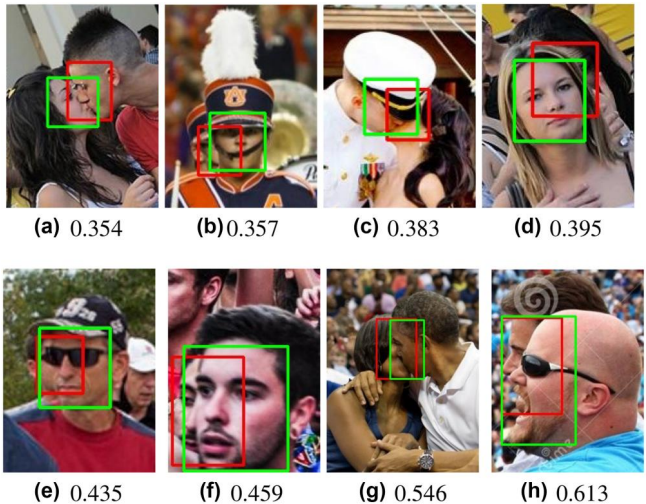

**(a)** 0.354  **(b)** 0.357  **(c)** 0.383  **(d)** 0.395
**(e)** 0.435  **(f)** 0.459  **(g)** 0.546  **(h)** 0.613

**FIGURE 5** Matching IoU threshold $T_m$ selection for overlap faces

**TABLE 3** Different IoU calibration interval $[T_m, T_c]$ for the Bounding-Box Deep Calibration (BDC) method

| Method | $[T_m, T_c]$ | Easy | Medium | Hard | Recall |
|---|---|---|---|---|---|
| Baseline | ∅ | 0.959 | 0.953 | 0.926 | 0.916 |
| Baseline + BDC | [0.5, 0.6] | 0.960 | 0.953 | 0.926 | 0.916 |
|  | [0.5, 0.7] | 0.961 | 0.954 | 0.926 | 0.916 |
|  | **[0.5, 0.8]** | **0.963** | **0.956** | **0.928** | **0.917** |
|  | [0.5, 0.9] | 0.962 | 0.954 | 0.925 | 0.916 |
|  | [0.5, 1.0] | 0.961 | 0.952 | 0.924 | 0.915 |

*Note*: The bold values indicate the best detection performance and its calibration interval of BDC. * Trained by the same face detector TinaFace.

our BDC algorithm, which can maintain the high recall rate of detection models. Besides, the IoU calibration interval of the baseline is considered as emptyset ∅. The performance under different calibration interval is listed in Table 3. After multiple ablative experiments, we find the optimal hyper-parameters. When IoU calibration interval $[T_m, T_c]$ is set to [0.5, 0.8], baseline with our proposed BDC method that can increase the average precision (AP) of 0.4% (Easy), 0.3% (Medium), 0.2% (Hard) and the recall rate of 0.1%. The results in Table 3 demonstrate the effectiveness of our proposed BDC method on improving models' precision and recall rate.

### 4.3.2 | The effectiveness of BDC on multiple heavy-weight detectors

We firstly train local detection models with original annotations, denoted as SSH(L), SFA(L), PyramidBox(L) and TinaFace(L). It should be noticed that PyramidBox[3] is a pytorch version while TinaFace is a simplified version offered by the

[3] https://github.com/Goingqs/PyramidBox

**TABLE 4** Effectiveness of Bounding-Box Deep Calibration (BDC) method on multiple heavy-weight detectors

| Detector | BDC | Easy | Medium | Hard |
|---|---|---|---|---|
| SSH(L) |  | 0.931 | 0.921 | 0.845 |
| SSH(B) | ✓ | 0.940 (+0.009) | 0.928 (+0.007) | 0.853 (+0.008) |
| SFA(L) |  | 0.949 | 0.936 | 0.866 |
| SFA(B) | ✓ | 0.949 (+0.000) | 0.937 (+0.001) | 0.873 (+0.007) |
| PyramidBox(L) |  | 0.953 | 0.943 | 0.892 |
| PyramidBox(B) | ✓ | 0.955 (+0.002) | 0.946 (+0.003) | 0.897 (+0.005) |
| TinaFace(L) |  | 0.959 | 0.953 | 0.926 |
| TinaFace(B) | ✓ | 0.963 (+0.004) | 0.956 (+0.003) | 0.928 (+0.002) |

*Note*: * All calibrated annotations on training set are generated by BDC method.

author. Then, we directly use the same calibrated annotations, as described in subsection 4.2.2 and subsection 4.2.3, to train models on multiple detectors, marked as SSH(B), SFA(B), PyramidBox(B) and TinaFace(B). The AP values of multiple heavy-weight detectors are listed in Table 4. As can be seen, SSH(B), SFA(B), PyramidBox(B) and TinaFace(B) consistently achieve the better AP performance than SSH(L), SFA(L), PyramidBox(L) and TinaFace(L), respectively. These results demonstrate the effectiveness of the BDC method on multiple detectors. We believe that our proposed BDC method can further improve the detection performance under the same experimental conditions.

### 4.3.3 | The effectiveness of BDC on multiple light-weight detectors

We further evaluate the effect of the BDC method on two light-weight detectors including RetinaFace [3] and YOLO5-Face [49]. RetinaFace utilises MobileNet-0.25 [50] as its backbone while YOLOv5n adopts ShuffleNetv2 [51]. Similar to heavy-weight detectors, we first train local detection models with original annotations, denoted as RetinaFace(L), YOLOv5n(L). Then, we directly use the same calibrated annotations, as described in subsection 4.2.2 and subsection 4.2.3, to train models on these two light-weight detectors, marked as RetinaFace(B), YOLOv5n(B). The evaluation results of multiple light-weight detectors are listed in Table 5. It should be noticed that RetinaFace has two modes to evaluate performance on the WIDER FACE validation set. Different from the fast mode with feature pyramid, the accurate mode also uses bounding-box voting. As can be seen, RetinaFace(B) and YOLOv5n(B) consistently achieve better AP performance and recall rate than RetinaFace(L) and YOLOv5n(L) by a large margin. These results demonstrate the effectiveness of the BDC method on multiple light-weight detectors. Besides, our simple and effective BDC method can improve face detection without adding extra inference time and memory consumption, which is more fit for light-weight detectors in real-time situations.

**TABLE 5** Effectiveness of Bounding-Box Deep Calibration (BDC) method on light-weight detectors

| Detector | BDC | Backbone | Mode | Easy | Medium | Hard | Recall |
|---|---|---|---|---|---|---|---|
| RetinaFace(L) | | MobileNet-0.25 | Fast | 0.847 | 0.829 | 0.726 | 0.787 |
| RetinaFace(B) | ✓ | MobileNet-0.25 | Fast | 0.850 (+0.003) | 0.842 (+0.013) | 0.751 (+0.025) | 0.801 (+0.014) |
| RetinaFace(L) | | MobileNet-0.25 | Accurate | 0.898 | 0.870 | 0.758 | 0.822 |
| RetinaFace(B) | ✓ | MobileNet-0.25 | Accurate | 0.907 (+0.009) | 0.884 (+0.014) | 0.782 (+0.024) | 0.840 (+0.018) |
| YOLOv5n(L) | | ShuffleNetv2 | – | 0.936 | 0.915 | 0.805 | – |
| YOLOv5n(B) | ✓ | ShuffleNetv2 | – | 0.943 (+0.007) | 0.922 (+0.007) | 0.814 (+0.009) | – |

*Note*: * All calibrated annotations on training set are generated by BDC method.

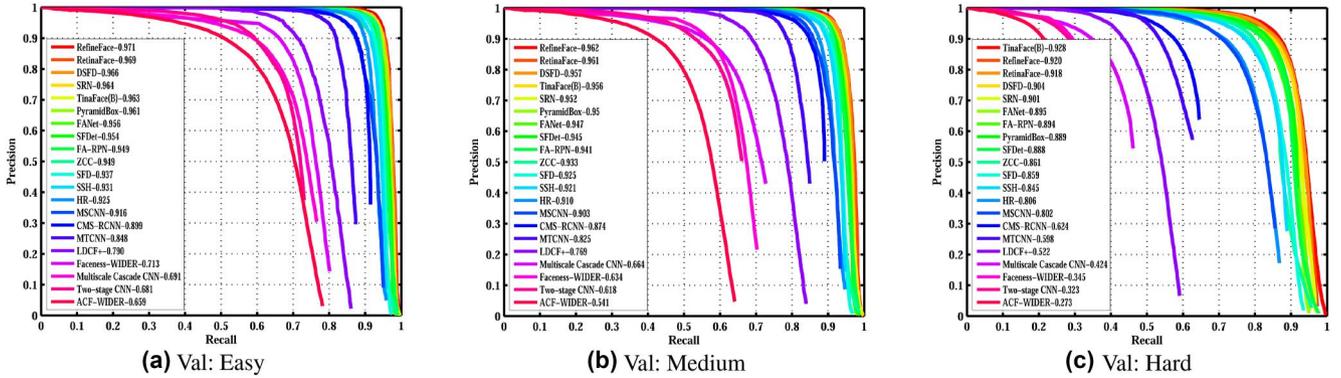

**FIGURE 6** Precision-recall curves on the WIDER FACE validation set

## 4.4 | Evaluation on benchmark

### 4.4.1 | WIDER FACE

We utilise calibrated annotations, generated by our proposed BDC method, to train the detection model TinaFace(B) and evaluate it on the WIDER FACE validation set against the recently published state-of-the-art face detection methods including RefineFace [42], RetinaFace [3], DSFD [36], SRN [39], PyramidBox [33], FANet [29], SFDet [52], FA-RPN [34], ZCC [32], $S^3$FD [35], SSH [41], HR [27], MSCNN [53], CMS-RCNN [54], MTCNN [26], LDCF+ [55], Faceness [56], Multiscale Cascade CNN [21], ACF [57] and Two-stage CNN [21]. Figure 6 shows the precision-recall curves and AP values on the WIDER FACE validation set. As seen in Figure 6, the performance gap among these algorithms is very small. Together with the results in Table 4 and Table 5, we believe BDC is a promising method to further improve detection performance on the WIDER FACE dataset.

### 4.4.2 | FDDB

We directly use the same detection model TinaFace(B) to perform the evaluation on the FDDB dataset. Specifically, the shortest side of the input images is set to 400 pixels while the larger side is less than 800 pixels. We compare TinaFace(B) against the recently published state-of-the-art methods including FANet [29], PyramidBox [33], DSFD [36], FD-CNN [58], ICC-CNN [59], RSA [60], $S^3$FD [35], FaceBoxes [61], HR [27], HR-ER [27], DeepIR [62], LDCF+ [55], UnitBox [43], Conv3D [63], Face Faster RCNN [64] and MTCNN [26] on the FDDB dataset. For a more fair comparison, the predicted bounding boxes are converted to bounding ellipses. Figure 7 shows the discrete ROC curves and continuous ROC curves of these methods on the FDDB dataset respectively. As can be seen, TinaFace(B) consistently achieves a relatively higher performance in terms of both the discrete ROC curves and continuous ROC curves. These results demonstrate the effectiveness and impressive generalisation capability of our proposed BDC method for improving performance in face detection.

## 5 | CONCLUSIONS AND FUTURE WORK

In this paper, we examined the misaligned results with high detection confidence but low localization accuracy on the training set and identified that misaligned annotations is the main reason. For the sake of further improving detection performance, a novel BDC method was proposed to reasonably replace misaligned annotations with model predicted bounding-boxes and create a new annotation file for training set. Extensive experiments demonstrate the effectiveness of the BDC method on improving models' precision and recall rate, without adding extra inference time and memory consumption. Our proposed method could be a general strategy for improving the performance of CNN-based face detectors, especially for light-weight detectors in real-time situations. To our knowledge, it is the first time for improving face detection to calibrate misaligned bounding-boxes on the training set.

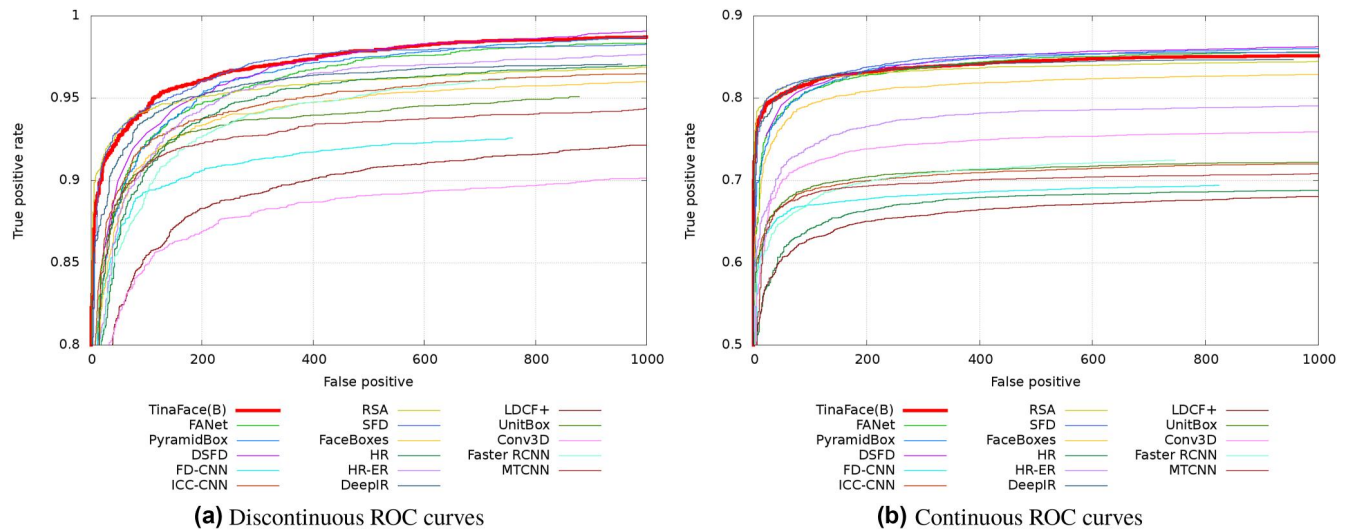

**FIGURE 7** Evaluation on the FDDB dataset

In the future, we attempt to alleviate annotation misalignment from the LDL perspective. Thus, annotation calibration may be extended to other computer vision tasks. Moreover, it would be an interesting direction to investigate the effect of calibration in terms of the labelling quantity and quality trade-off.

### ACKNOWLEDGEMENTS
This work was supported in part by the National Natural Science Foundation of China (No. 61801190), the National Key Research and Development Project of China (No. 2019YFC0409105), the Industrial Technology Research and Development Funds of Jilin Province (No. 2019C054-3), the 'Thirteenth Five-Year Pla' Scientific Research Planning Project of Education Department of Jilin Province (Nos. JKH20200678KJ and JJKH20200997KJ), the Fundamental Research Funds for the Central Universities, JLU, and in part by the Graduate Innovation Fund of Jilin University.

### CONFLICT OF INTEREST
The authors have no relevant conflicts of interest to disclose.

### DATA AVAILABILITY STATEMENT
Research data are not shared.

### ORCID
*Shi Luo* https://orcid.org/0000-0001-5334-0453
*Xiaoli Zhang* https://orcid.org/0000-0001-8412-4956